# Optimization of Real, Hermitian Quadratic Forms:

## Real, Complex Hopfield-Amari Neural Network

Dr. G. Rama Murthy PhD
Associate Professor
IIIT Hyderabad
Andhra Pradesh, India
E-mail: rammurthy@iiit.ac.in

B. Nischal(B.Tech)
NIT Hamirpur
Himachal Pradesh, India
E-mail: nishi4dadd@gmail.com

*Abstract*—In this research paper, the problem of optimization of quadratic forms associated with the dynamics of Hopfield-Amari neural network is considered. An elegant (and short) proof of the states at which local/global minima of quadratic form are attained is provided. A theorem associated with local/global minimization of quadratic energy function using the Hopfield-Amari neural network is discussed. The results are generalized to a "Complex Hopfield neural network" dynamics over the complex hypercube (using a "complex signum function"). It is also reasoned through two theorems that there is no loss of generality in assuming the threshold vector to be a zero vector in the case of real as well as a "Complex Hopfield neural network". Some structured quadratic forms like Toeplitz form and Complex Toeplitz form are discussed.

## I. INTRODUCTION

In natural, engineering and other sciences as well as technology dynamical systems are utilized for modeling the associated dynamic phenomena. Large body of literature is developed for analyzing and synthesizing linear as well as non-linear dynamical systems. Amari as well as Hopfield successfully discovered as well as formalized an artificial neural network modeling an associative memory. In the following the operation of such a neural network is briefly discussed for completeness:

This neural network can be represented as *N(W,T)* where *W* signifies the synaptic weight matrix and T the threshold vector. The order of such a network is defined as the number of nodes in the network and denoted by n. T is the threshold vector consisting of the individual thresholds attached to each of the nodes of the neural network, $W_{ij}$ represents the synaptic weight from the $j_{th}$ node to the $i_{th}$ node and $V_i(t)$ represents the state of node *i* at time *t*. The network can be operated in two modes- If the state computation is done at one particular node at any instant of time, then the network is said to be in serial mode of operation and if the state computation is done at all the nodes at any instant of time, then the network is said to be in fully parallel mode of operation. The consequent state computation of any arbitrary node (*i*) is as follows:

$$V_i(t+1) = Sgn(H_i(t)) = \quad 1, \text{if } H_i(t) \geq 0$$
$$= \quad -1, \text{otherwise}$$

where $H_i(t) = \sum_{j=1}^{n} W_{i,j} V_j(t) - T_i$,

A state $V(t)$ is said to be stable if and only if $V(t) = Sgn(WV(t))$. We define an anti-stable state as follows- A state $V(t)$ is said to be anti-stable if and only if $V(t) = -Sgn(WV(t))$.

There is a convergence theorem given by Hopfield associated with local/global maximization of quadratic energy function and is as follows- Let *N*=(*W, T*) be a neural network, with *W* being a synaptic weight matrix which is a symmetric matrix with nonnegative diagonal elements and *T* being the threshold vector; then the following hold.

*1)* If N is operating in a serial mode, then the network will always converge to a stable state.

*2)* If N is operating in a fully parallel mode, then the network will always converge to a stable state or to a cycle of length 2.

In contrast to the above problem, the authors naturally formulated the problem of minimization of real quadratic form over the real hypercube. The authors attempted and succeeded in providing an elegant and short proof of the condition satisfied by the local/global minimum anti-stable state on the real hypercube. This proof is provided in Section II. In this section we also discuss local/global minimization of real quadratic form based energy function over the real hypercube using the Hopfield-Amari neural network. In this section we also reason that there is no loss of generality in choosing the energy function of a Hopfield-Amari neural network as a quadratic form. In Section III minimization of quadratic forms associated with a Hermitian matrix (i.e., Hermitian forms) over the complex hypercube using the complex signum function is discussed. The elegant proof in Section II is generalized in the complex case. Local/global minimization of Hermitian form based energy function over the complex hypercube using the complex Hopfield-Amari neural network is also discussed. In Section III we also reason that there is no loss of generality in choosing the energy function of a complex Hopfield-Amari neural network as a Hermitian form. In Section IV the generalizations of the Toeplitz as well as the complex Toeplitz forms are provided.

We expect several problems in engineering, computer science, natural sciences and other areas requiring the minimization of quadratic form (real or Hermitian form) over the real /complex hypercube. One important motivation for the problem is the characterization of corner positive (corner positive definite) matrices that arise in the characterization of autocorrelation function of {+1, -1} valued random processes.

Definition: A real valued matrix B is corner positive if and only if $X^T B X \geq 0$ for all $X$ lying on the symmetric, real hypercube.

## II. Optimization Of Real Quadratic Forms Over The Real Hypercube

### A. Theorem(1) Statement

Let E be an arbitrary n × n real matrix. If u minimizes the quadratic form, $x^T E x$ subject to the constraint $|u_i| = 1$ for $1 \leq i \leq n$ (i.e., u lies on the real hypercube), then

$$u = -\text{sign}(Cu), \quad (1)$$

where C is the symmetric matrix with zero diagonal elements obtained from E.

### B. Proof

Without loss of generality any arbitrary real matrix can be broken into a symmetric as well as an anti-symmetric matrix.

$$E = \frac{E + E^T}{2} + \frac{E - E^T}{2}$$

Let,

$$P = \frac{E + E^T}{2}, \quad Q = \frac{E - E^T}{2}$$

The quadratic form associated with Q is always zero where as the quadratic form associated with P can be written as

$$X^T P X = \sum_{i=1}^{n} \sum_{j=1(i=j)}^{n} p_{ij}\, x_i x_j + \sum_{i=1}^{n} \sum_{j=1(i \neq j)}^{n} p_{ij}\, x_i x_j$$

$$= \sum_{i=1}^{n} p_{ii} x_i^2 + \sum_{i=1}^{n} \sum_{j=1(i \neq j)}^{n} p_{ij}\, x_i x_j$$

If X lies on the real hypercube, $|x_i| = 1$.
Therefore,

$$X^T P X = \text{Trace}(P) + \sum_{i=1}^{n} \sum_{j=1(i \neq j)}^{n} p_{ij}\, x_i x_j \quad (2)$$

In view of (2), without loss of generality, C may be assumed to be a symmetric matrix with zero diagonal elements obtained from E. Now we claim that a binary vector, which provides the minimum value for the quadratic form necessarily satisfies (1).

Suppose it does not. This implies that there exists a vector **v**, which achieves the minimum of the quadratic form and at the same time violates the condition (1). Suppose all the other components of **v** satisfy the condition (1) except the nth component. (If it is not the case, then just permute the rows and columns).

$$v_n = \text{sign}(\sum_{j=1}^{n} c_{nj}\, v_j)$$

Define a new vector **w** from **v** in the following manner

$$w_j = v_j \text{ if } j \neq n \text{ and } w_n = -v_n$$

Now consider the quadratic form evaluated with **w**

$$[\widetilde{w}\ :\ w_n]^T \begin{bmatrix} D & b \\ b^T & 0 \end{bmatrix} \begin{bmatrix} \widetilde{w} \\ w_n \end{bmatrix}$$
$$= \widetilde{w}^T D\, \widetilde{w} + 2\, \widetilde{w}^T b w_n$$

Defining $v = [v^T\colon v_n]$ and subtracting the quadratic form value with **v,** we have

$$w^T C w - v^T C v = -4\, v^T b\, v_n$$

Now on using the fact that the nth component of **v** violates the condition (1), we have that

$$w^T C w < v^T C v$$

Thus we have the desired contradiction.

*Remark (1):* It can be noted that the same proof can actually be used to prove the condition associated with maximization.

### C. Theorem(2)

Now that we provided a characterization of anti-stable states (local minima of a real quadratic form over a real hypercube), we are naturally led to the question how they can be reached starting from an arbitrary initial state. The following theorem provides a method of reaching the local/global minima of a real quadratic form based energy function (with the threshold vector being non-zero) over the real hypercube using the Hopfield-Amari neural network. It is a convergence theorem in the spirit of the related theorem for Hopfield-Amari neural network [1].

*Statement*

Let $N=(W, T)$ be a neural network, with $W$ being the synaptic weight matrix which is a symmetric matrix with nonnegative diagonal elements and T being the threshold vector; then the following hold.

*1)* If N is operating in a serial mode and the consequent state computation of any arbitrary node (*i*) is as follows:
$$V_i(t+1) = -sign(H_i(t)) = -1, \text{if } H_i(t) \geq 0$$
$$= 1, \text{otherwise}$$
where $H_i(t) = \sum_{j=1}^{n} W_{i,j}\, V_j(t) - T_i$,
then the network will always converge to an anti-stable state.

*2)* If N is operating in a fully parallel mode and the consequent state computation of any arbitrary node (*i*) is similar to that of the serial mode, then the network will always converge to an anti-stable state or to a cycle of length 2.

*Proof*

The proof follows from considering the minimization of the energy function (Lyapunov function) defined on the dynamics of Hopfield-Amari neural network. The generalization follows from a similar argument.

*D. Theorem(3)*

The theorem (2) holds true even for the real quadratic form, as it can be proved that there is no loss of generality in neglecting the threshold vector.

*Statement*

There is no loss of generality in neglecting the threshold vector in a Hopfield-Amari neural network i.e., there is no loss of generality in choosing the energy function of a Hopfield-Amari neural network as a quadratic form.

*Proof*

In the following, we reason that in a Hopfield network there is no loss of generality in assuming that the threshold at each neuron is zero.

Consider a Hopfield network uniquely defined by (*W,T*) where *W* is the synaptic weight matrix which is a symmetric matrix with nonnegative diagonal elements and *T* is the threshold vector.

Now, we consider a new Hopfield network with a dummy node i.e., the $(n+1)^{th}$ node. Let the state of the $(n+1)^{th}$ node be fixed at '1'. Also the synaptic weights from $(n+1)^{th}$ node to the $i^{th}$ node are chosen as $-T_i$ (i.e., the $i^{th}$ component of the vector *T*). The last diagonal element $w_{n+1,n+1}$ is suitably chosen to ensure that the state at the $(n+1)^{th}$ node is always fixed at 1. Thus the synaptic weight matrix of the resulting Hopfield network is given by-

$$\widetilde{W} = \begin{bmatrix} W & -T \\ -T^T & k \end{bmatrix}, \text{ where } k = \sum |T_i| + 1.$$

Also $\widetilde{T} = 0$ i.e., the threshold vector is a zero vector. It is now easy to reason that the serial and parallel modes of operation of the new Hopfield network retain the structure of stable/anti-stable states of the original network.

Thus, the energy function of the new Hopfield network is a quadratic form (since the threshold vector is zero).

*E. Associative Memory Application*

We know that the Hopfield-Amari neural network can be used as an associative memory. Hopfield synthesized a real valued synaptic weight matrix from the patterns to be stored in such a way that the network so obtained has these patterns as stable states. The weight matrix given by Hopfield is as follows:

$$W = \sum_{j=1}^{S}(X_j X_j^T - I)$$

where $S$ is the number of patterns to be stored, $I$ is the identity matrix and $X_1, X_2 \ldots X_S$ are the real patterns (orthogonal to each other and also lying on the real hypercube) to be stored.

We synthesized a real valued synaptic weight matrix from the patterns to be stored in such a way that the network so obtained has these patterns as anti-stable states. The weight matrix is as follows:

$$W = \sum_{j=1}^{S}(-X_j X_j^T + I)$$

where $S$ is the number of patterns to be stored, $I$ is the identity matrix and $X_1, X_2 \ldots X_S$ are the real patterns (orthogonal to each other and also lying on the real hypercube) to be stored.

*Note:* The real patterns $X_1, X_2 \ldots X_S$ are said to be orthogonal to each other if $X_j^T X_k = 0 \ \forall \ j \neq k \ (1 \leq j \leq s, 1 \leq k \leq s)$ and $X_j^T X_j = 1 \ \forall \ j \ (1 \leq j \leq s)$.

III. OPTIMIZATION OF HERMITIAN FORMS OVER THE COMPLEX HYPERCUBE

In the theory of complex quadratic forms, there are various efforts made on studying Hermitian forms. To the best of our knowledge, no one studied Hermitian forms and their minimization over the complex hypercube. So, now in the following discussion we provide precise definition of the underlying concepts involved:

*A. Complex Hypercube*

Unlike the real hypercube there are $4^n$ points on the complex hypercube of n-dimensions as a vector of size n on the complex hypercube has each entry belonging to the set:

{1+j, 1-j, -1+j, -1-j}.

*B. Complex Signum Function*

We define the complex signum function as follows:
Sgn(a+jb) = sign(a) + j sign(b)

*C. Equations*

The following equations related to arbitrary complex quadratic forms (not just Hermitian forms) are here by provided for completeness. Specifically, their evaluation on the complex hypercube is one of our contributions.

Any complex valued matrix, A can be decomposed in the following manner:

$$A = \frac{A + A^\star}{2} + \frac{A - A^\star}{2}$$

$$= A_H + A_{SH}$$

Where $A_H$ is the Hermitian part associated with A and $A_{SH}$ is the skew-Hermitian part. These parts have the following properties:

$$A_H = A_H^\star, \; a_{ji} = a_{ij}^\star$$

$$A_{SH} = -A_{SH}^\star, \; a_{ji} = -a_{ij}^\star$$

Now, let us evaluate the quadratic form associated with A.

$$X^\star A X = X^\star A_H X + X^\star A_{SH} X$$

$$X^\star A_H X = \sum_i \sum_{j(i=j)} x_i^\star a_H(i,j) x_j + \sum_i \sum_{j(i \neq j)} x_i^\star a_H(i,j) x_j$$

$$= \sum_i a_H(i,i) |x_i|^2 + \sum_i \sum_{j(i \neq j)} x_i^\star a_H(i,j) x_j$$

The sum,

$\sum_i a_H(i,i) |x_i|^2$ is real, since $a_H(i,i)$ is real and $|x_i|^2$ is real.

*1) Remark:* In the case of a complex hypercube $|x_i| = \sqrt{2}$ Hence, $\sum_i a_H(i,i) |x_i|^2 = 2\sum_i a_H(i,i) = 2 \text{ Trace}(A_H)$. Therefore,
$X^\star A_H X = 2 \text{ Trace}(A_H) + \sum_i \sum_{j(i \neq j)} x_i^\star a_H(i,j) x_j$ \quad (3)

Considering the other sum, $\sum_i \sum_{j(i \neq j)} x_i^\star a_H(i,j) x_j$, this contains the sum of terms of the form,

$S = x_i^\star A_H(i,j) x_j + x_j^\star A_H(j,i) x_i$. Since, $A_H(j,i) = A_H^\star(i,j)$

$S = x_i^\star A_H(i,j) x_j + (x_i^\star A_H(i,j) x_j)^\star = 2Re(x_i^\star A_H(i,j) x_j)$. Thus the net contribution of all such terms is real. Hence the quadratic form associated with a Hermitian matrix is always real.

*Note:* If the analysis is carried out for a skew-Hermitian matrix in the similar way, we can conclude that the quadratic form associated with a skew-Hermitian matrix is always imaginary.

*D. Theorem(4) Statement*

Let E be an arbitrary n × n Hermitian matrix. If u minimizes the Hermitian form, $x^\star E x$ subject to the constraint that u lies on the complex hypercube, then

u = -Sgn(Cu), \quad (4)

where C is the matrix with zero diagonal elements obtained from E.

*E. Proof*

In view of (3), without loss of generality, C may be assumed to be a matrix with zero diagonal elements obtained from E.

Hence, C can be written as follows, $C = \begin{bmatrix} D & b \\ b^\star & 0 \end{bmatrix}$

Let us assume that there exists a vector w such that it minimizes the Hermitian form $x^\star E x$ and also satisfies (4).

$$w = -Sgn(Cw)$$

Defining w as $[\widetilde{w}: w_n]$ and evaluating $w^\star C w$ we get

$$w^\star C w = [\widetilde{w}^\star \; w_n^\star] \begin{bmatrix} D & b \\ b^\star & 0 \end{bmatrix} \begin{bmatrix} \widetilde{w} \\ w_n \end{bmatrix}$$
$$= \widetilde{w}^\star D \widetilde{w} + \widetilde{w}^\star b w_n + w_n^\star b^\star \widetilde{w}$$
$$= \widetilde{w}^\star D \widetilde{w} + 2Re(w_n^\star b^\star \widetilde{w})$$

Also, we have $\begin{bmatrix} \widetilde{w} \\ w_n \end{bmatrix} = -Sgn(\begin{bmatrix} D & b \\ b^\star & 0 \end{bmatrix} \begin{bmatrix} \widetilde{w} \\ w_n \end{bmatrix})$

$$\Rightarrow w_n = -Sgn(b^\star \widetilde{w})$$

If we can prove that there exists no v such that it minimizes the Hermitian form $x^\star E x$ and at the same time violates the condition (4), then the theorem (4) can be proved.

We just vary the nth component of v in order to achieve this.

Let $v = [\widetilde{w}: v_n]$. Now three possibilities arise for $v_n$.

$v_n = -w_n$ or $w_n^\star$ or $-w_n^\star$

$w^\star C w - v^\star C v = 2Re(w_n^\star b^\star \widetilde{w}) - 2Re(v_n^\star b^\star \widetilde{w})$

Let $b^\star \widetilde{w} = p + jq \Rightarrow w_n = -sign(p) - jsign(q)$

$\Rightarrow w_n^\star = -sign(p) + jsign(q)$

This implies,

$v_n^\star = -sign(p) - jsign(q)$ or $sign(p) + jsign(q)$ or $sign(p) - jsign(q)$

Ruling out all the three cases,

Case (1):

$v_n^\star = -sign(p) - jsign(q)$,

$\Rightarrow w^\star C w - v^\star C v = -4 \, qsign(q)$

$qsign(q) > 0 \Rightarrow w^\star C w < v^\star C v$.

Case (2):

$v_n^\star = sign(p) + jsign(q)$,

$\Rightarrow w^\star C w - v^\star C v = -4 \, psign(p)$

$psign(p) > 0 \Rightarrow w^\star C w < v^\star C v$.

Case (3):

$v_n^\star = sign(p) - jsign(q)$,

$\Rightarrow w^\star C w - v^\star C v = -4 \, psign(p) - 4 \, qsign(q)$

$psign(p) + qsign(q) > 0 \Rightarrow w^\star C w < v^\star C v$.

Thus all the three cases are ruled out and hence there exists no v such that it minimizes the Hermitian form $x^\star E x$ and at the same time violates the condition (4). Hence the theorem (4) is proved.

*Remark (2):* It can be noted that the same proof can actually be used to prove the condition associated with maximization.

### F. Theorem(5)

Now that we provided a characterization of anti-stable states (local minima of a Hermitian form over a complex hypercube), we are naturally led to the question how they can be reached starting from an arbitrary initial state. The following theorem provides a method of reaching the local/global minima of a Hermitian form based energy function (with the threshold vector being non-zero) over the complex hypercube using the complex Hopfield-Amari neural network. It is a convergence theorem in the spirit of the related theorem for complex Hopfield-Amari neural network [2].

*Statement*

Let $N=(W, T)$ be a neural network, with $W$ being a synaptic weight matrix which is a Hermitian matrix with nonnegative diagonal elements and $T$ being the threshold vector; then the following hold.

*1)* If N is operating in a serial mode and the consequent state computation of any arbitrary node (*i*) is as follows:
$V_i(t+1) = -Sgn(H_i(t))$, where
$H_i(t) = \sum_{j=1}^{n} W_{i,j} V_j(t) - T_i$,
then the network will always converge to an anti-stable state.

*2)* If N is operating in a fully parallel mode and the consequent state computation of any arbitrary node (*i*) is similar to that of the serial mode, then the network will always converge to an anti-stable state or to a cycle of length 2.

*Proof*

The proof follows from considering the minimization of the energy function defined on the dynamics of complex Hopfield-Amari neural network. The generalization follows from a similar argument.

### G. Theorem(6)

The theorem (5) holds true even for the Hermitian form, as it can be proved that there is no loss of generality in neglecting the threshold vector.

*Statement*

There is no loss of generality in neglecting the threshold vector in a complex Hopfield-Amari neural network i.e., there is no loss of generality in choosing the energy function of a complex Hopfield-Amari neural network as a Hermitian form.

*Proof*

In the following, we reason that in a complex Hopfield network there is no loss of generality in assuming that the threshold at each neuron is zero.

Consider a complex Hopfield network uniquely defined by $(W,T)$ where $W$ is the synaptic weight matrix which is a Hermitian matrix and $T$ is the threshold vector.

Now, we consider a new Hopfield network with a dummy node i.e., the $(n+1)^{th}$ node. Let the state of the $(n+1)^{th}$ node be fixed at '$1+j$'. Also the synaptic weights from $(n+1)^{th}$ node to the $i^{th}$ node ($S_i$) are chosen in such a way that $(1+j) \times S_i = T_i$ (i.e., the $i^{th}$ component of the vector $T$).

Let us suppose $T_i = a_i + jb_i \ \forall \ i \ (1 \leq i \leq n)$ then we choose $S_i = c_i + jd_i \ \forall \ i \ (1 \leq i \leq n)$ in such a way that $(c_i + jd_i) \times (1+j) = a_i + jb_i$. Thus, the set-
$c_i - d_i = a_i$ and $c_i + d_i = b_i$ can be solved as follows:

$$\begin{bmatrix} b_i \\ a_i \end{bmatrix} = \begin{bmatrix} 1 & 1 \\ 1 & -1 \end{bmatrix} \begin{bmatrix} c_i \\ d_i \end{bmatrix} \text{ i.e., } \begin{bmatrix} b_i \\ a_i \end{bmatrix} = H_2 \begin{bmatrix} c_i \\ d_i \end{bmatrix},$$

where $H_2$ is a Hadamard matrix of order 2.

$$\Rightarrow H_2^{-1} \begin{bmatrix} b_i \\ a_i \end{bmatrix} = \begin{bmatrix} c_i \\ d_i \end{bmatrix} = \begin{bmatrix} \tfrac{1}{2} & \tfrac{1}{2} \\ \tfrac{1}{2} & -\tfrac{1}{2} \end{bmatrix} \begin{bmatrix} b_i \\ a_i \end{bmatrix}$$

The last diagonal element $w_{n+1,n+1}$ is suitably chosen to ensure that the state at the $(n+1)^{th}$ node is always fixed at '$1+j$'. Thus the synaptic weight matrix of the resulting Hopfield network is given by-

$$\widetilde{W} = \begin{bmatrix} W & -S \\ -S^\star & k \end{bmatrix}, \text{ where } k = (\sum |Re(T_i)| + |Im(T_i)|) + 1.$$

Also $\widetilde{T} = 0$ i.e., the threshold vector is a zero vector. It is now easy to reason that the serial and parallel modes of operation of the new complex Hopfield network retain the structure of stable/anti-stable states of the original network.

Thus, the energy function of the new complex Hopfield network is a Hermitian form (since the threshold vector is zero).

### H. Associative Memory Application

We synthesized a complex valued Hermitian synaptic weight matrix from the patterns to be stored in such a way that the network so obtained has these patterns as stable states. The weight matrix is as follows:

$$W = \sum_{j=1}^{S} (X_j X_j^\star - 2I)$$

where $S$ is the number of patterns to be stored, $I$ is the identity matrix and $X_1, X_2 \ldots X_S$ are the complex patterns (unitary to each other and also lying on the complex hypercube) to be stored.

We synthesized a complex valued Hermitian synaptic weight matrix from the patterns to be stored in such a way that the network so obtained has these patterns as anti-stable states. The weight matrix is as follows:

$$W = \sum_{j=1}^{S}(-X_j X_j^\star + 2I)$$

where $S$ is the number of patterns to be stored, $I$ is the identity matrix and $X_1, X_2 \ldots X_S$ are the complex patterns (unitary to each other and also lying on the complex hypercube) to be stored.

*Note:* The complex patterns $X_1, X_2 \ldots X_S$ are said to be unitary to each other if $X_j^\star X_k = 0 \;\forall\; j \neq k \;(1 \leq j \leq s, 1 \leq k \leq s)$ and $X_j^\star X_j = 1 \;\forall\; j \;(1 \leq j \leq s)$.

### IV. SOME STRUCTURED QUADRATIC FORMS OVER THE HYPERCUBE

#### A. Toeplitz form

We define a Toeplitz form as follows- A quadratic form is said to be Toeplitz if it is of the form $X^T T X$, where $T$ is a symmetric Toeplitz matrix. The beauty of a symmetric Toeplitz matrix is if the first row of the matrix is defined, the entire matrix can be defined. Toeplitz form is called a structured quadratic form as it can be generalized as follows:

$$X^T T X = \sum_{i=1}^{n} x_i^2 T_{11} + 2\sum_{k=2}^{n} T_{1k} \sum_{\substack{i=1 \\ j=i+k-1, j\leq n}}^{n} x_i x_j$$

where $T$ is an $n \times n$ matrix.

Considering this Toeplitz form over the real hypercube, we get

$$X^T T X = nT_{11} + 2\sum_{k=2}^{n} T_{1k} \sum_{\substack{i=1 \\ j=i+k-1, j\leq n}}^{n} x_i x_j$$

#### B. Complex Toeplitz form

We define a complex Toeplitz form as follows- A quadratic form is said to be complex Toeplitz if it is of the form $X^\star T X$, where $T$ is a Hermitian Toeplitz matrix. The Hermitian Toeplitz matrix possesses the same property as that of a Toeplitz matrix and that is if the first row of the matrix is defined, the entire matrix can be defined. The complex Toeplitz form is also a structured quadratic form and can be generalized as follows:

$$X^\star T X = \sum_{i=1}^{n} x_i^\star T_{11} x_i + \sum_{k=2}^{n} T_{1k} \sum_{\substack{i=1 \\ j=i+k-1, j\leq n}}^{n} x_i^\star x_j + \sum_{k=2}^{n} T_{1k}^\star \sum_{\substack{i=1 \\ j=i+k-1, j\leq n}}^{n} x_i x_j^\star$$

where $T$ is an $n \times n$ matrix.

Considering this complex Toeplitz form over the complex hypercube, we get

$$X^T T X = 2nT_{11} + \sum_{k=2}^{n} T_{1k} \sum_{\substack{i=1 \\ j=i+k-1, j\leq n}}^{n} x_i^\star x_j + \sum_{k=2}^{n} T_{1k}^\star \sum_{\substack{i=1 \\ j=i+k-1, j\leq n}}^{n} x_i x_j^\star$$